\pdfoutput=1

\documentclass[11pt]{article}

\usepackage{emnlp2021}

\usepackage{times}
\usepackage{latexsym}

\usepackage[T1]{fontenc}

\usepackage[utf8]{inputenc}

\usepackage{microtype}

\usepackage{times}
\usepackage{latexsym}
\usepackage{graphicx}
\usepackage{subcaption}
\usepackage{enumitem}
\usepackage{multicol}
\usepackage{multirow}
\usepackage{makecell}
\usepackage{amsmath}
\usepackage{algorithm}
\usepackage{algorithmic}

\title{A Framework for Rationale Extraction for Deep QA models}

\author{Sahana Ramnath\thanks{* now at Google Research, Bangalore}, Preksha Nema\footnotemark[1], Deep Sahni, Mitesh M. Khapra  \\
Robert Bosch Centre for Data Science and AI (RBC-DSAI) \\ Indian Institute of Technology Madras, Chennai, India \\ 
\texttt{\{sahanjich,preksha.nema9,deep.sahani11\}@gmail.com,} \\ \texttt{miteshk@cse.iitm.ac.in}}

\begin{document}
\maketitle
\begin{abstract}
As neural-network-based QA models become deeper and more complex, there is a demand for robust frameworks which can access a model's rationale for its prediction. Current techniques that provide insights on a model's working are either dependent on adversarial datasets or are proposing models with explicit explanation generation components. These techniques are time-consuming and challenging to extend to existing models and new datasets. In this work, we use  `Integrated Gradients' to extract rationale for existing state-of-the-art models in the task of Reading Comprehension based Question Answering (RCQA). On detailed analysis and comparison with collected human rationales, we find that though $\sim$40-80\% words of extracted rationale coincide with the human rationale (precision), only 6-19\% of human rationale is present in the extracted rationale (recall).
\end{abstract}

\section{Introduction}
There has been a surge of increasingly complex neural architectures, with the primary focus of increasing the model's performance on one or more specific RCQA datasets. Indeed, the development of such complex models has lead to improved performance. However, as the models' predictive capacity increases, it becomes necessary to understand and evaluate \emph{how} the model resulted in the given prediction. 

One line of works  ~\citep{si2019does,Jia:17} analyzes models on adversarial datasets and provides insights into whether such models can understand the passage and question correctly. The others \citep{lei2016rationalizing} introduce and train a rationale generation/extraction component as a part of the model itself. However, creating an adversarial dataset that could highlight (most of/all) the model's limitations is an expensive and difficult task. Similarly, having a rationale generation component in the model is not a technique that can be extended to analyze existing works.

\begin{table}
    \small
    \begin{tabular}{p{7cm}}
        \Xhline{3\arrayrulewidth}
        \textbf{Question:} What color pants did the Broncos wear in Super Bowl 50 ?\\
        \textbf{Answer:} white\\
         \hline 
         \textbf{Rationale Extracted for DCN} \\
            \begin{tabular}{p{7cm}}
                \textcolor{blue}{\textit{As the designated home team in the annual rotation between AFC}} and \textcolor{blue}{\textit{NFC teams}} , the \textcolor{blue}{\textit{Broncos}} elected to wear their road \textcolor{blue}{\textit{white}} jerseys with matching white pants. Elway stated, ``we have had Super Bowl success in our white uniforms." The Broncos last wore matching white jerseys and pants in the super bowl XXXII.
            \end{tabular}        \\
         \textbf{Rationale Extracted for BERT} \\
            \begin{tabular}{p{7cm}}
                As the designated home team in the annual rotation between AFC and NFC teams, the Broncos elected to wear their road white jerseys with matching white \textcolor{blue}{\textit{pants}}. Elway stated, ``we have had Super Bowl success in our white uniforms." The Broncos last wore matching white jerseys and \textcolor{blue}{\textit{pants}} in the super bowl XXXII.
            \end{tabular}        \\            
        \Xhline{3\arrayrulewidth}
        \end{tabular}  
    \caption{Sample of extracted rationale (italicized and highlighted in blue) for DCN and BERT using our proposed framework. Both models predicted the correct answer ``white''.}
    \label{tab:example}
\end{table}

\noindent Therefore in this work, we propose a simple yet effective rationale extraction framework that can be used to analyze any existing/new model. We first measure the importance of each passage word to the model's predicted answer using `Integrated Gradients'\citep{Sundararajan:17}. We further conduct detailed quantitative and qualitative analysis on four established deep QA models - BiDAF~\citep{Seo:16}, BERT~\citep{Devlin:18}, DCN~\citep{Xiong:16} and QANet~\citep{Yu:18} for the widely adopted SQuAD \citep{Rajpurkar:16} dataset. We then compare the extracted rationales to collected human rationales for 500 samples. We compute precision and recall for words in the model's rationale with the words in the human rationale. We find that while BERT is highly precise (80\%), it is likely because it highlights very few words as its rationale. DCN, on the other hand, has the highest recall of 0.18; however, such a low recall indicates very low completeness compared to human rationales. 

\noindent For example, in Table \ref{tab:example}, we can observe from DCN's rationale that the model focuses on ``white jerseys'' to reach the answer rather than ``white pants'', which shows the reasoning to get to the answer is not entirely correct. The extracted rationale for BERT is further incomplete; it just highlights ``pants'', giving no insight into how the answer has been predicted. This shows the need to evaluate rationales (explanations) extracted (generated) from deep models for a better understanding of how the model is predicting the answer. 

\section{Related Work}
A wide variety of Deep QA models ~\citep{Seo:16, Xiong:16, Yu:18} have been introduced to solve RCQA tasks across various datasets
~\citep{Rajpurkar:16,Lai:17,Nguyen:16,Joshi:17}. In particular, various BERT-based models like ~\citet{Liu:19,Lan:19,Devlin:18}  have approached human-level performance on these datasets. While these models show impressive performances, there are doubts about how they arrive at their predictions. Hence, the community is strongly advocating for more interpretable models. There have been several works dedicated to analyze the interpretability of simple encoder-decoder based neural models using various path-based attribution methods (Integrated Gradients \cite{Sundararajan:17}, Layer-wise Relevance Propagation \cite{Bach:15}, white-box methods like LIME \cite{Ribeiro:16}). Some works \citep{arras2017explaining,li2016understanding} analyze deep models for simpler tasks such as text classification, NLI, etc. However, so far, similar methods have not been used to analyze deeper and more complex RCQA models. In our work, we present a simple yet effective technique to adopt integrated gradients to extract one possible rationale behind a model's prediction. 

\section{Why Integrated Gradients?}
Among various attribution methods such as LRP \citep{Bach:15}, LIME \citep{Ribeiro:16}, DeepLift \citep{Shrikumar:17}, we use \textbf{Integrated Gradients} to extract rationales because of the following reason: As discussed in \citet{Sundararajan:17}, attribution methods should satisfy the following properties (i) adhere to sensitivity, \textit{i.e.}, they should focus only on relevant features and not on irrelevant features (ii) show implementation invariance, \textit{i.e.},  the calculated attributions should be identical for two functionally equivalent networks irrespective of the implementation. \citet{Sundararajan:17} argues that all the above methods except Integrated Gradients fail to satisfy at least one of these properties. For example, attribution methods such as LRP \citep{Bach:15} and DeepLift \citep{Shrikumar:17} replace gradients with discrete gradients and use a modified form of backpropagation. Hence, these methods depend heavily on the implementation used for the discrete gradients. Further, they require different implementations for different model architectures, and will thus render our analysis incomparable. 
Further, various works \citep{Serrano:19,jain2019attention} have demonstrated that in-built attention mechanisms do not accurately reflect the model's attribution over input items. The word embeddings could change uninterpretably from the input to the attention layer. Hence, at best, attention scores could represent attribution \textit{at} the attention layer, and not layers below it.    

\section{Analysis on Extracted Rationale}
To evaluate a model's explainability, it is first essential to define it. We adopt this definition from \citet{serrano2019attention}: \textit{a model's explanation is the minimal set of items (in this case, words) which when removed collectively, causes a change in the model's prediction.}. To enumerate over the subsets of passage words feasibly, we pick the top items as ranked by integrated gradients. We describe below how we chose the minimal subset or \emph{indicator} words.

\subsection{Framework for Rationale Extraction}
\noindent For a given passage $D$ with $d$ words $[w_1,w_2,\dots,w_d]$, question $Q$, answer $[i,j]$, where $i$ and $j$ are start and end indices of answer span in $D$, and a model $M$ with parameters $\phi$, the answer prediction task is modeled as:
\begin{align}
    \nonumber p(w_s, w_e) &= M(w_s, w_e|\text{embed}(D),\text{embed}(Q),\phi)
\end{align}
 $w_s,w_e$ are the start and end indices of the predicted answer span. The passage and question words are embedded using $\text{embed}(.)$, that encodes a word to $\mathbf{R}^L$ space.

We can now compute Integrated Gradient for a passage word $w_i$, embedded as $x_i \in \mathbf{R}^L$, as follows:
\begin{align}
\nonumber IG(x_i) =  \int\limits_{\alpha=0}^1\frac{\partial M(\tilde{x} + \alpha(x_i - \tilde{x}))}{\partial x_i}\  \ d\alpha 
\end{align}
where $\tilde{x}$ is a zero vector, that serves as a baseline to measure integrated gradient for $w_i$. We approximate the above integral across $50$ uniform samples between $[0,1]$.

\begin{algorithm}
\caption{Rationale Extraction}
\label{rationale-extraction}
\begin{algorithmic}[1]
\STATE $\tilde{D}([w_1,...,w_d]) \gets||IG(w_1)||,...,||IG(w_d)||$
\STATE // normalize for a probability distribution
\STATE $\tilde{D} \gets \tilde{D}/\text{sum}(\tilde{D})$ 
\STATE $\mathbf{X} \leftarrow $ Rank $[w_1,\dots,w_d]$ based on $\tilde{D}$
\STATE $\text{indicator\_words} = [\ ]$
\REPEAT
\STATE $\text{indicator\_words}\text{.insert(pop}(\mathbf{X}))$
\UNTIL{(Decision Flips)}
\end{algorithmic}
\end{algorithm}
\noindent {Finally we use Algorithm \ref{rationale-extraction}, to extract rationale. We compute the $IG(x_i)$ for all passage words $i \in [1,d]$. We further compute importance scores for all $w_i$'s by taking the euclidean norm of $IG(w_i)$, which is then normalized to a probability distribution $\tilde{D}$. We start removing the most important words from the passage (and replace it with $\mathbf{0}$), until the model predicts a wrong answer, i.e., when there is a \emph{decision flip} (Line 6-8).}

\subsection{Quantitative Analysis on Rationale Extracted}
Using the definition for model explainability, we can say a model is \textit{more} explainable if a smaller set of words causes its decision to flip. To quantify the model's explainability as per this description, we define the measure \textit{flip fraction}:  the total fraction of words in the passage that constitute the model's rationale. That is, a model is more explainable if the flip-fraction is lower.
We compute these flip fractions for the entire dev split of SQuAD\footnote{10k samples, each with a 100-300 words passage, a natural language query and answer span in the passage}. The mean and variance of these fractions across all samples are calculated and depicted in Table \ref{tab:decflip_stat_all}.

\begin{table}[h]
    \centering
    \begin{tabular}{ccc}
        \hline
        \textbf{Model} & \textbf{Mean} & \textbf{Variance} \\
        \hline
        BERT & 0.072 & 0.03 \\
        BiDAF & 0.161 & 0.064 \\
        QANet & 0.165 & 0.065 \\
        DCN & 0.215 & 0.059 \\
        \hline
    \end{tabular}
    \caption{Mean and Variance of flip-fraction across SQuAD's dev set, scale of 0-1}
    \label{tab:decflip_stat_all}
\end{table}
\noindent
We see that all four models have low flip-fractions for the majority of the dataset. This means that decision flips happen as soon as the most important words (as ranked by the model) are removed. From this, we can conclude that the model is able to highlight what is important for its prediction. That is, it is able to attribute its important features correctly (regardless of how logical this explanation is). 

\noindent A potential limitation with the flip fraction metric is that a lower score does not necessarily ensure the \textit{quality} of the rationale. Even with a low fraction, the model might be selecting wrong/insufficient words. Specifically, for the case of QA where the answer is a span from the passage, the model could directly drop the answer span itself, making it difficult to understand how exactly the model reached the answer without focusing on other necessary words. Hence, in the next section, we compare the extracted rationale to the rationale annotated by humans.

\begin{table*}
    \small
    \begin{tabular}{p{15cm}}
        \Xhline{3\arrayrulewidth}
        \textbf{Question:}What was Maria Curie the first female recipient of ?\\
        \textbf{Answer:} Nobel Prize\\
         \hline 
             \begin{tabular}{p{1cm} p{13cm}}
                \textbf{DCN} & \textcolor{blue}{\textit{One of the most famous people born in Warsaw was Maria}} SkÃÂodowska-Curie, who achieved international recognition for her research on radioactivity and was the first \textcolor{blue}{\textit{female}} recipient of the Nobel Prize....
            \end{tabular}        \\
            \begin{tabular}{p{1cm}p{13cm}}
                \textbf{BiDAF} & One of the most famous people born in Warsaw was Maria SkÃÂodowska-Curie, who achieved international recognition for her research on radioactivity and was the first \textcolor{blue}{ \textit{female recipient of the Nobel}} Prize....
            \end{tabular}        \\
            \begin{tabular}{p{1cm}p{13cm}}
                \textbf{QANET} & One of the most famous people born in Warsaw was \textcolor{blue}{\textit{Maria SkÃÂodowska-Curie}}, who achieved international recognition for her research on radioactivity and was the first \textcolor{blue}{\textit{female}} recipient of the \textcolor{blue}{\textit{Nobel Prize}}....
            \end{tabular}        \\
            \begin{tabular}{p{1cm}p{13cm}}
                \textbf{BERT} & One of the most famous people born in Warsaw was Maria SkÃÂodowska-Curie, who achieved international recognition for her research on radioactivity and was the first female recipient of the \textcolor{blue}{\textit{Nobel Prize}}....
            \end{tabular}        \\            
        \Xhline{3\arrayrulewidth}
        \end{tabular}  
    \caption{Sample of Rationale Extracted for DCN, BiDAF, QANET and BERT using our proposed framework. The rationale comprises of the italicized/blue words. All the models predicted the correct answer ``Nobel Prize''.}.
    \label{tab:full-anlysis}
\end{table*}

\subsection{Human Evaluations on Rationale Extracted}
To analyze the quality of the extracted rationale, we picked 500 datapoints randomly and equally split amongst different question types\footnote{what, where, which, who, when, why, how}. For the corresponding \{\textit{passage, query, answer}\} triplets, we asked two human annotators to mark words and phrases in the passage, which they think are the most important to arrive at the answer. We explicitly asked them not to mark \textit{only} the answer. We then paired up two annotators and asked them to resolve any discrepancies in what they individually thought was the rationale. We removed points for which the annotators did not reach a consensus. Using these human annotations as baseline, we check the model's rationale for \textit{precision} and \textit{recall (completeness)}. As defined in Algorithm \ref{rationale-extraction}, the model's rationale is the set of indicator words at the model's input layer. The model is defined as \textit{precise} if every word in its rationale is also present in the human-annotated rationale. It is defined as \textit{complete} if every word in the human rationale is present in its rationale. While comparing the two sets, we do not take stop words into account. \\
\begin{table}[]
\centering
\resizebox{\linewidth}{!}{%
\begin{tabular}{|c|c|c|c|c|c|c|}
\hline
\multirow{2}{*}{\textbf{Model}} & \multicolumn{3}{c|}{\textbf{Incl. Answer Span}} & \multicolumn{3}{c|}{\textbf{Excl. Answer Span}} \\ \cline{2-7} 
 & \textbf{\% P} & \textbf{\% R} & \textbf{\% F1} & \textbf{\% P} & \textbf{\% R} & \textbf{\% F1} \\ \hline
\textbf{BERT} & 94.9 & 17.45 & 29.1 & 22.8 & 5.1 & 8.3 \\ \hline
\textbf{BiDAF} & 85.8 & 19.8 & 32.2 & 29.4 & 8.7 &  13.4 \\ \hline
\textbf{DCN} & 65.1 & 26.9 & 38.1 & 22.7 & 14.4 &  17.6\\ \hline
\textbf{QANet} & 83.1 & 19.6 & 31.7 & 28.3 & 8.2 & 12.7 \\ \hline
\end{tabular}%
}
\caption{Precision(P), Recall(R) and F1 score of overlap of models' extracted rationale with human rationale}
\label{tab:human_rationale}
\end{table}



\noindent Our preliminary analysis found that the models often mark their predicted answer span as part of their rationale. Hence, we compare the models' rationale to the human annotation as two cases: (i) including the answer span, which may be present in none/either/both of them (ii) excluding the answer span. The results can be found in Table \ref{tab:human_rationale}.\\

On analysis, we find that while precision is reasonably high in both cases, all the models are highly focused on the answer span itself; on excluding the answer, there is approximately a 10-20\% drop in precision for all models. Further, we observe that the recall values are much lower; this concludes that though the models seemed explainable based on flip fractions, the reasoning on how the answer was retrieved is not yet sufficient. \\
We observe that DCN's recall (though low in an absolute sense) is the highest across the four models. We hypothesize that this is because the modeling of query-passage interaction is a simple extension of cosine similarity compared to the highly non-linear functions in other models.

\noindent \textbf{Qualitative Analysis:} We further show rationales extracted from all four models for one \textit{\{passage, question, answer\}} triplet in Table \ref{tab:full-anlysis}. The human rationale was ``maria, curie, first female recipient of noble prize''.  We infer that DCN's rationale contains necessary words such as `maria', `female', and `of' required to link the question to the answer. However, it still misses out on necessary words like ``first, recipient'' to complete the rationale. Moreover, it highlights words like `most famous people born' which are irrelevant to answer the question, and thus has lower precision. BiDAF and QANET, do not highlight any unnecessary words and are almost complete. BiDAF fails to highlight ``Maria, first'' and QANET does not highlight ``first, recipient''. On the other hand, BERT does not give any information in the rationale; it just selects the answer word directly. This shows that though BERT is the best-performing model on SQuAD, with the existing attribution methods, it does not provide meaningful rationales. 

\section{Conclusion}
In this work, we proposed a framework to extract rationales for RCQA models. Based on one definition of model explainability (from \citet{serrano2019attention}), we saw that all four models could be deemed explainable. However, based on human evaluation and qualitative analysis, we found that though the models are comparable to humans in terms of performance, there is a wide gap in terms of the rationale given for its prediction.  In the future, we would like to extend this framework to datasets other than SQuAD and analyze other QA models.

\section*{Acknowledgements}
We thank the Department of Computer Science and Engineering, IIT Madras and the Robert Bosch Center for  Data  Science  and  Artificial  Intelligence, IIT Madras (RBC-DSAI) for providing us compute resources. We thank Google for supporting Preksha Nema contribution through the Google Ph.D. Fellowship programme.

\bibliography{custom}
\bibliographystyle{acl_natbib}

\end{document}